# Randomized Algorithms for the Loop Cutset Problem

**Ann Becker**                                           ANYUTA@CS.TECHNION.AC.IL
**Reuven Bar-Yehuda**                                    REUVEN@CS.TECHNION.AC.IL
**Dan Geiger**                                           DANG@CS.TECHNION.AC.IL
*Computer Science Department*
*Technion, Haifa, 32000, Israel*

## Abstract

We show how to find a minimum weight loop cutset in a Bayesian network with high probability. Finding such a loop cutset is the first step in the method of conditioning for inference. Our randomized algorithm for finding a loop cutset outputs a minimum loop cutset after $O(c\,6^k kn)$ steps with probability at least $1 - (1 - \frac{1}{6^k})^{c6^k}$, where $c > 1$ is a constant specified by the user, $k$ is the minimal size of a minimum weight loop cutset, and $n$ is the number of vertices. We also show empirically that a variant of this algorithm often finds a loop cutset that is closer to the minimum weight loop cutset than the ones found by the best deterministic algorithms known.

## 1. Introduction

The method of conditioning is a well known inference method for the computation of posterior probabilities in general Bayesian networks (Pearl, 1986, 1988; Suermondt & Cooper, 1990; Peot & Shachter, 1991) as well as for finding MAP values and solving constraint satisfaction problems (Dechter, 1999). This method has two conceptual phases. First to find an optimal or close to optimal loop cutset and then to perform a likelihood computation for each instance of the variables in the loop cutset. This method is routinely used by geneticists via several genetic linkage programs (Ott, 1991; Lang, 1997; Becker, Geiger, & Schaffer, 1998). A variant of this method was developed by Lange and Elston (1975).

Finding a minimum weight loop cutset is NP-complete and thus heuristic methods have often been applied to find a reasonable loop cutset (Suermondt & Cooper, 1990). Most methods in the past had no guarantee of performance and performed very badly when presented with an appropriate example. Becker and Geiger (1994, 1996) offered an algorithm that finds a loop cutset for which the logarithm of the state space is guaranteed to be at most a constant factor off the optimal value. An adaptation of these approximation algorithms has been included in version 4.0 of FASTLINK, a popular software for analyzing large pedigrees with small number of genetic markers (Becker et al., 1998). Similar algorithms in the context of undirected graphs are described by Bafna, Berman, and Fujito (1995) and Fujito (1996).

While approximation algorithms for the loop cutset problem are quite useful, it is still worthwhile to invest in finding a minimum loop cutset rather than an approximation because the cost of finding such a loop cutset is amortized over the many iterations of the conditioning method. In fact, one may invest an effort of complexity exponential in the size of the loop cutset in finding a minimum weight loop cutset because the second phase of the conditioning algorithm, which is repeated for many iterations, uses a procedure of such complexity. The same considerations apply also to constraint satisfaction problems as well as other problems in which the method of conditioning is useful (Dechter, 1990, 1999).

In this paper we describe several randomized algorithms that compute a loop cutset. As done by Bar-Yehuda, Geiger, Naor, and Roth (1994), our solution is based on a reduction to the weighted feedback vertex set problem. A *feedback vertex set (FVS)* $F$ is a set of vertices of an undirected graph $G = (V, E)$ such that by removing $F$ from $G$, along with all the edges incident with $F$, a set of trees is obtained. The *Weighted Feedback Vertex Set (WFVS) problem* is to find a feedback vertex set $F$ of a vertex-weighted graph with a weight function $w : V \rightarrow \mathbb{R}^+$, such that $\sum_{v \in F} w(v)$ is minimized. When $w(v) \equiv 1$, this problem is called the FVS problem. The decision version associated with the FVS problem is known to be NP-Complete (Garey & Johnson, 1979, pp. 191–192).

Our randomized algorithm for finding a WFVS, called REPEATEDWGUESSI, outputs a minimum weight FVS after $O(c\, 6^k kn)$ steps with probability at least $1 - (1 - \frac{1}{6^k})^{c6^k}$, where $c > 1$ is a constant specified by the user, $k$ is the minimal size of a minimum weight FVS, and $n$ is the number of vertices. For unweighted graphs we present an algorithm that finds a minimum FVS of a graph $G$ after $O(c\, 4^k kn)$ steps with probability at least $1 - (1 - \frac{1}{4^k})^{c4^k}$. In comparison, several deterministic algorithms for finding a minimum FVS are described in the literature. One has a complexity $O((2k + 1)^k n^2)$ (Downey & Fellows, 1995b) and others have a complexity $O((17k^4)! n)$ (Bodlaender, 1990; Downey & Fellows, 1995a).

A final variant of our randomized algorithms, called WRA, has the best performance because it utilizes information from previous runs. This algorithm is harder to analyze and its investigation is mostly experimental. We show empirically that the actual run time of WRA is comparable to a Modified Greedy Algorithm (MGA), described by Becker and Geiger (1996), which is the best available deterministic algorithm for finding close to optimal loop cutsets, and yet, the output of WRA is often closer to the minimum weight loop cutest than the output of MGA.

The rest of the paper is organized as follows. In Section 2 we outline the method of conditioning, explain the related loop cutset problem and describe the reduction from the loop cutset problem to the WFVS Problem. In Section 3 we present three randomized algorithms for the WFVS problem and their analysis. In Section 4 we compare experimentally WRA and MGA with respect to output quality and run time.

## 2. Background: The Loop Cutset Problem

A short overview of the method of conditioning and definitions related to Bayesian networks are given below. See the book by Pearl (1988) for more details. We then define the loop cutset problem.

Let $P(u_1, \ldots, u_n)$ be a probability distribution where each variable $u_i$ has a finite set of possible values called the *domain of* $u_i$. A directed graph $D$ with no directed cycles is called a *Bayesian network of $P$* if there is a 1–1 mapping between $\{u_1, \ldots, u_n\}$ and vertices in $D$, such that $u_i$ is associated with vertex $i$ and $P$ can be written as follows:

$$P(u_1, \ldots, u_n) = \prod_{i=1}^{n} P(u_i \mid u_{i_1}, \ldots, u_{i_{j(i)}}) \qquad (1)$$

where $i_1, \ldots, i_{j(i)}$ are the source vertices of the incoming edges to vertex $i$ in $D$.

Suppose now that some variables $\{v_1, \ldots, v_l\}$ among $\{u_1, \ldots, u_n\}$ are assigned specific values $\{\mathbf{v}_1, \ldots, \mathbf{v}_l\}$ respectively. The *updating problem* is to compute the probability $P(u_i \mid v_1 = \mathbf{v}_1, \ldots, v_l = \mathbf{v}_l)$ for $i = 1, \ldots, n$.

A *trail* in a Bayesian network is a subgraph whose underlying graph is a simple path. A vertex $b$ is called a *sink* with respect to a trail $t$ if there exist two consecutive edges $a \rightarrow b$ and $b \leftarrow c$ on $t$. A trail $t$ is *active by a set of vertices $Z$* if (1) every sink with respect to $t$ either is in $Z$ or has a descendant in $Z$ and (2) every other vertex along $t$ is outside $Z$. Otherwise, the trail is said to be *blocked (d-separated)* by $Z$.

Verma and Pearl proved that if $D$ is a Bayesian network of $P(u_1, \ldots, u_n)$ and all trails between a vertex in $\{r_1, \ldots, r_l\}$ and a vertex in $\{s_1, \ldots, s_k\}$ are blocked by $\{t_1, \ldots, t_m\}$, then the corresponding sets of variables $\{u_{r_1}, \ldots, u_{r_l}\}$ and $\{u_{s_1}, \ldots, u_{s_k}\}$ are independent conditioned on $\{u_{t_1}, \ldots, u_{t_m}\}$ (Verma & Pearl, 1988). Furthermore, Geiger and Pearl proved that this result cannot be enhanced (Geiger & Pearl, 1990). Both results were presented and extended by Geiger, Verma, and Pearl (1990).

Using the close relationship between blocked trails and conditional independence, Kim and Pearl developed an algorithm UPDATE-TREE that solves the updating problem on Bayesian networks in which every two vertices are connected with at most one trail (Kim & Pearl, 1983). Pearl then solved the updating problem on any Bayesian network as follows (Pearl, 1986). First, a set of vertices $S$ is selected such that any two vertices in the network are connected by at most one *active* trail in $S \cup Z$, where $Z$ is any subset of vertices. Then, UPDATE-TREE is applied once for each combination of value assignments to the variables corresponding to $S$, and, finally, the results are combined. This algorithm is called the method of *conditioning* and its complexity grows exponentially with the size of $S$. The set $S$ is called a *loop cutset*. Note that when the domain size of the variables varies, then UPDATE-TREE is called a number of times equal to the product of the domain sizes of the variables whose corresponding vertices participate in the loop cutset. If we take the logarithm of the domain size (number of values) as the weight of a vertex, then finding a loop cutset such that the sum of its vertices weights is minimum optimizes Pearl's updating algorithm in the case where the domain sizes may vary.

We now give an alternative definition for a loop cutset $S$ and then provide a probabilistic algorithm for finding it. This definition is borrowed from a paper by Bar-Yehuda et al. (1994). The underlying graph $G$ of a directed graph $D$ is the undirected graph formed by ignoring the directions of the edges in $D$. A *cycle* in $G$ is a path whose two terminal vertices coincide. A *loop* in $D$ is a subgraph of $D$ whose underlying graph is a cycle. A vertex $v$ is a *sink* with respect to a loop $\Gamma$ if the two edges adjacent to $v$ in $\Gamma$ are directed into $v$. Every loop must contain at least one vertex that is not a sink with respect to that loop. Each vertex that is not a sink with respect to a loop $\Gamma$ is called an *allowed vertex with respect to $\Gamma$*. A *loop cutset* of a directed graph $D$ is a set of vertices that contains at least one allowed vertex with respect to each loop in $D$. The weight of a set of vertices $X$ is denoted by $w(X)$ and is equal to $\sum_{v \in X} w(v)$ where $w(x) = \log(|x|)$ and $|x|$ is the size of the domain associated with vertex $x$. A *minimum weight loop cutset* of a weighted directed graph $D$ is a loop cutset $F^*$ of $D$ for which $w(F^*)$ is minimum over all loop cutsets of $G$. The *Loop Cutset Problem* is defined as finding a minimum weight loop cutset of a given weighted directed graph $D$.

The approach we take is to reduce the loop cutset problem to the weighted feedback vertex set problem, as done by Bar-Yehuda et al. (1994). We now define the weighted feedback vertex set problem and then the reduction.

Let $G = (V, E)$ be an undirected graph, and let $w : V \to I\!R^+$ be a weight function on the vertices of $G$. A *feedback vertex set* of $G$ is a subset of vertices $F \subseteq V$ such that each cycle in $G$ passes through at least one vertex in $F$. In other words, a feedback vertex set $F$ is a set of vertices of $G$ such that by removing $F$ from $G$, along with all the edges incident with $F$, we obtain a set of trees (i.e., a forest). The weight of a set of vertices $X$ is denoted (as before) by $w(X)$ and is equal to $\sum_{v \in X} w(v)$. A *minimum feedback vertex set* of a weighted graph $G$ with a weight function $w$ is a feedback vertex set $F^*$ of $G$ for which $w(F^*)$ is minimum over all feedback vertex sets of $G$. The *Weighted Feedback Vertex Set (WFVS) Problem* is defined as finding a minimum feedback vertex set of a given weighted graph $G$ having a weight function $w$.

The reduction is as follows. Given a weighted directed graph $(D, w)$ (e.g., a Bayesian network), we define the *splitting* weighted undirected graph $D_s$ with a weight function $w_s$ as follows. Split each vertex $v$ in $D$ into two vertices $v_{\text{in}}$ and $v_{\text{out}}$ in $D_s$ such that all incoming edges to $v$ in $D$ become undirected incident edges with $v_{\text{in}}$ in $D_s$, and all outgoing edges from $v$ in $D$ become undirected incident edges with $v_{\text{out}}$ in $D_s$. In addition, connect $v_{\text{in}}$ and $v_{\text{out}}$ in $D_s$ by an undirected edge. Now set $w_s(v_{\text{in}}) = \infty$ and $w_s(v_{\text{out}}) = w(v)$. For a set of vertices $X$ in $D_s$, we define $\psi(X)$ as the set obtained by replacing each vertex $v_{\text{in}}$ or $v_{\text{out}}$ in $X$ by the respective vertex $v$ in $D$ from which these vertices originated. Note that if $X$ is a cycle in $D_s$, then $\psi(X)$ is a loop in $D$, and if $Y$ is a loop in $D$, then $\psi^{-1}(Y) = \bigcup_{v \in Y} \psi^{-1}(v)$ is a cycle in $D_s$ where

$$\psi^{-1}(v) = \begin{cases} v_{\text{in}} & v \text{ is a sink on } Y \\ v_{\text{out}} & v \text{ is a source on } Y \\ \{v_{\text{in}}, v_{\text{out}}\} & \text{otherwise} \end{cases}$$

(A vertex $v$ is a *source* with respect to a loop $Y$ if the two edges adjacent to $v$ in $Y$ originate from $v$). This mapping between loops in $D$ and cycles in $D_s$ is one-to-one and onto.

Our algorithm can now be easily stated.

## ALGORITHM LoopCutset

**Input:** *A Bayesian network $D$*
**Output:** *A loop cutset of $D$*
    1. Construct the splitting graph $D_s$
        with weight function $w_s$
    2. Find a feedback vertex set $F$ for $(D_s, w_s)$
        using the Weighted Randomized Algorithm (WRA)
    3. Output $\psi(F)$.

It is immediately seen that if WRA (developed in later sections) outputs a feedback vertex set $F$ of $D_s$ whose weight is minimum with high probability, then $\psi(F)$ is a loop cutset of $D$ with minimum weight with the same probability. This observation holds due to the one-to-one and onto correspondence between loops in $D$ and cycles in $D_s$ and because WRA never chooses a vertex that has an infinite weight.

## 3. Algorithms for the WFVS Problem

Recall that a feedback vertex set of $G$ is a subset of vertices $F \subseteq V$ such that each cycle in $G$ passes through at least one vertex in $F$. In Section 3.1 we address the problem of finding a FVS with a minimum number of vertices and in Sections 3.2 and 3.3 we address the problem of finding a FVS with a minimum weight. Throughout, we allow $G$ to have parallel edges. If two vertices $u$ and $v$ have parallel edges between them, then every FVS of $G$ includes either $u$, $v$, or both.

### 3.1 The Basic Algorithms

In this section we present a randomized algorithm for the FVS problem. First we introduce some additional terminology and notation. Let $G = (V, E)$ be an undirected graph. The degree of a vertex $v$ in $G$, denoted by $d(v)$, is the number of vertices adjacent to $v$. A *self-loop* is an edge with two endpoints at the same vertex. A *leaf* is a vertex with degree less or equal 1, a *linkpoint* is a vertex with degree 2 and a *branchpoint* is a vertex with degree strictly higher than 2. The cardinality of a set $X$ is denoted by $|X|$.

A graph is called *rich* if every vertex is a branchpoint and it has no self-loops. Given a graph $G$, by repeatedly removing all leaves, and bypassing with an edge every linkpoint, a graph $G'$ is obtained such that the size of a minimum FVS in $G'$ and in $G$ are equal and every minimum FVS of $G'$ is a minimum FVS of $G$. Since every vertex involved in a self-loop belongs to every FVS, we can transform $G'$ to a rich graph $G^r$ by adding the vertices involved in self loops to the output of the algorithm.

Our algorithm is based on the observation that if we pick an edge at random from a rich graph there is a probability of at least $1/2$ that at least one endpoint of the edge belongs to any given FVS $F$. A precise formulation of this claim is given by Lemma 1 whose proof is given implicitly by Voss (1968, Lemma 4).

**Lemma 1** *Let $G = (V, E)$ be a rich graph, $F$ be a feedback vertex set of $G$ and $X = V \setminus F$. Let $E_X$ denote the set of edges in $E$ whose endpoints are all vertices in $X$ and $E_{F,X}$ denote the set of edges in $G$ that connect vertices in $F$ with vertices in $X$. Then, $|E_X| \leq |E_{F,X}|$.*

**Proof.** The graph obtained by deleting a feedback vertex set $F$ of a graph $G(V, E)$ is a forest with vertices $X = V \setminus F$. Hence, $|E_X| < |X|$. However, each vertex in $X$ is a branchpoint in $G$, and so,

$$3|X| \leq \sum_{v \in X} d(v) = |E_{F,X}| + 2|E_X|.$$

Thus, $|E_X| \leq |E_{F,X}|$. □

Lemma 1 implies that when picking an edge at random from a rich graph, it is at least as likely to pick an edge in $E_{F,X}$ than an edge in $E_X$. Consequently, selecting a vertex at random from a randomly selected edge has a probability of at least $1/4$ to belong to a minimum FVS. This idea yields a simple algorithm to find a FVS.

**ALGORITHM SingleGuess(G,j)**

**Input:** *An undirected graph $G_0$ and an integer $j > 0$.*

**Output:** *A feedback vertex set $F$ of size $\leq j$, or "Fail" otherwise.*

    **For** $i = 1, \ldots, j$

        1. Reduce $G_{i-1}$ to a rich graph $G_i$
            while placing self loop vertices in $F$.

        2. **If** $G_i$ is the empty graph **Return** $F$

        3. Pick an edge $e = (u, v)$ at random from $E_i$

        4. Pick a vertex $v_i$ at random from $(u, v)$

        5. $F \leftarrow F \cup \{v_i\}$

        6. $V \leftarrow V \setminus \{v_i\}$

    **Return** "Fail"

Due to Lemma 1, when $\textsc{SingleGuess}(G, j)$ terminates with a FVS of size $j$, there is a probability of at least $1/4^j$ that the output is a minimum FVS.

Note that steps 3 and 4 in $\textsc{SingleGuess}$ determine a vertex $v$ by first selecting an arbitrary edge and then selecting an arbitrary endpoint of this edge. An equivalent way of achieving the same selection rule is to choose a vertex with probability proportional to its degree:

$$p(v) = \frac{d(v)}{\sum_{u \in V} d(u)} = \frac{d(v)}{2 \cdot |E|}$$

To see the equivalence of these two selection methods, define $\Gamma(v)$ to be a set of edges whose one endpoint is $v$, and note that for graphs without self-loops,

$$p(v) = \sum_{e \in \Gamma(v)} p(v|e) \cdot p(e) = \frac{1}{2} \sum_{e \in \Gamma(v)} p(e) = \frac{d(v)}{2 \cdot |E|}$$

This equivalent phrasing of the selection criterion is easier to extend to the weighted case and will be used in the following sections.

An algorithm for finding a minimum FVS with high probability, which we call $\textsc{Repeated Guess}$, can now be described as follows: Start with $j = 1$. Repeat $\textsc{SingleGuess}$ $c\,4^j$ times where $c > 1$ is a parameter defined by the user. If in one of the iterations a FVS of size $\leq j$ is found, then output this FVS, otherwise, increase $j$ by one and continue.

**ALGORITHM RepeatedGuess(G,c)**

**Input:** *An undirected graph $G$*
           *and a constant $c > 1$.*

**Output:** *A feedback vertex set $F$.*

    **For** $j = 1, \ldots, |V|$

        **Repeat** $c\,4^j$ **times**

            1. $F \leftarrow \textsc{SingleGuess}(G, j)$

            2. **If** $F$ is not "Fail" **then Return** $F$

        **End** {Repeat}

    **End** {For}

The main claims about these algorithms are given by the following theorem.

**Theorem 2** *Let $G$ be an undirected graph and $c \geq 1$ be a constant. Then, SINGLEGUESS$(G, k)$ outputs a FVS whose expected size is no more than $4k$, and REPEATEDGUESS$(G, c)$ outputs, after $O(c\,4^k kn)$ steps, a minimum FVS with probability at least $1 - (1 - \frac{1}{4^k})^{c4^k}$, where $k$ is the size of a minimum FVS and $n$ is the number of vertices.*

The claims about the probability of success and number of steps follow immediately from the fact that the probability of success of SINGLEGUESS$(G, j)$ is at least $(1/4)^j$ and that, in case of success, $O(c\,4^j)$ iterations are performed each taking $O(jn)$ steps. The result follows from the fact that $\sum_{j=1}^{k} j\,4^j$ is of order $O(k4^k)$. The proof about the expected size of a single guess is presented in the next section.

Theorem 2 shows that each guess produces a FVS which, on the average, is not too far from the minimum, and that after enough iterations, the algorithm converges to the minimum with high probability. In the weighted case, discussed next, we managed to achieve each of these two guarantees in separate algorithms, but we were unable to achieve both guarantees in a single algorithm.

## 3.2 The Weighted Algorithms

We now turn to the weighted FVS problem (WFVS) of size $k$ which is to find a feedback vertex set $F$ of a vertex-weighted graph $(G, w)$, $w : V \to I\!R^+$, of size less or equal $k$ such that $w(F)$ is minimized.

Note that for the weighted FVS problem we cannot replace each linkpoint $v$ with an edge because if $v$ has weight lighter than its branchpoint neighbors then $v$ can participate in a minimum weight FVS of size $k$.

A graph is called *branchy* if it has no endpoints, no self loops, and, in addition, each linkpoint is connected only to branchpoints (Bar-Yehuda, Geiger, Naor, & Roth, 1994). Given a graph $G$, by repeatedly removing all leaves, and bypassing with an edge every linkpoint that has a neighbor with equal or lighter weight, a graph $G'$ is obtained such that the weight of a minimum weight FVS (of size $k$) in $G'$ and in $G$ are equal and every minimum WFVS of $G'$ is a minimum WFVS of $G$. Since every vertex with a self-loop belongs to every FVS, we can transform $G'$ to a branchy graph without self-loops by adding the vertices involved in self loops to the output of the algorithm.

To address the WFVS problem we offer two slight modifications to the algorithm SINGLEGUESS presented in the previous section. The first algorithm, which we call SINGLEWGUESSI, is identical to SINGLEGUESS except that in each iteration we make a reduction to a branchy graph instead of a reduction to a rich graph. It chooses a vertex with probability proportional to the degree using $p(v) = d(v)/\sum_{u \in V} d(u)$. Note that this probability does not take the weight of a vertex into account. A second algorithm, which we call SINGLEWGUESSII, chooses a vertex with probability proportional to the ratio of its degree over its weight,

$$p(v) = \frac{d(v)}{w(v)} \Big/ \sum_{u \in V} \frac{d(u)}{w(u)}. \tag{2}$$

## ALGORITHM SingleWGuessI(G,j)

**Input:** *An undirected weighted graph $G_0$*
*and an integer $j > 0$.*

**Output:** *A feedback vertex set $F$ of size $\leq j$,*
*or "Fail" otherwise.*

**For** $i = 1, \ldots, j$

 1. Reduce $G_{i-1}$ to a branchy graph $G_i(V_i, E_i)$
   while placing self loop vertices in $F$.

 2. **If** $G_i$ is the empty graph **Return** $F$

 3. Pick a vertex $v_i \in V_i$ at random with
   probability $p_i(v) = d_i(v)/\sum_{u \in V_i} d_i(u)$

 4. $F \leftarrow F \cup \{v_i\}$

 5. $V \leftarrow V \setminus \{v_i\}$

**Return** "Fail"

The second algorithm uses Eq. 2 for computing $p(v)$ in Line 3. These two algorithms have remarkably different guarantees of performance. Version I guarantees that choosing a vertex that belongs to any given FVS is larger than $1/6$, however, the expected weight of a FVS produced by version I cannot be bounded by a constant times the weight of a minimum WFVS. Version II guarantees that the expected weight of its output is bounded by 6 times the weight of a minimum WFVS, however, the probability of converging to a minimum after any fixed number of iterations can be arbitrarily small. We first demonstrate via an example the negative claims. The positive claims are phrased more precisely in Theorem 3 and proven thereafter.

Consider the graph shown in Figure 1 with three vertices $a$,$b$ and $c$, and corresponding weights $w(a) = 6$, $w(b) = 3\epsilon$ and $w(c) = 3m$, with three parallel edges between $a$ and $b$, and three parallel edges between $a$ and $c$. The minimum WFVS $F^*$ with size 1 consists of vertex $a$. According to Version II, the probability of choosing vertex $a$ is (Eq. 2):

$$p(a) = \frac{\epsilon}{(1 + 1/m) \cdot \epsilon + 1}$$

So if $\epsilon$ is arbitrarily small and $m$ is sufficiently large, then the probability of choosing vertex $a$ is arbitrarily small. Thus, the probability of choosing a vertex from some $F^*$ by the criterion $d(v)/w(v)$, as done by Version II, can be arbitrarily small. If, on the other hand, Version I is used, then the probability of choosing $a, b,$ or $c$ is $1/2, 1/4, 1/4$, respectively. Thus, the expected weight of the first vertex to be chosen is $3/4 \cdot (\epsilon + m + 4)$, while the weight of a minimum WFVS is 6. Consequently, if $m$ is sufficiently large, the expected weight of a WFVS found by Version I can be arbitrarily larger than a minimum WFVS.

The algorithm for repeated guesses, which we call RepeatedWGuessI$(G, c, j)$ is as follows: repeat SingleWGuessI$(G, j)$ $c\,6^j$ times, where $j$ is the minimal number of vertices of a minimum weight FVS we seek. If no FVS is found of size $\leq j$, the algorithm outputs that the size of a minimum WFVS is larger than $j$ with high probability, otherwise, it outputs the lightest FVS of size less or equal $j$ among those explored. The following theorem summarizes the main claims.

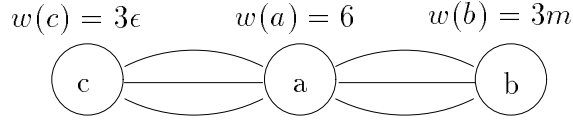

Figure 1: The minimum WFVS $F^* = \{a\}$.

**Theorem 3** *Let $G$ be a weighted undirected graph and $c \geq 1$ be a constant.*
*a) The algorithm* REPEATEDWGUESSI$(G, c, k)$ *outputs after* $O(c\,6^k kn)$ *steps a minimum WFVS with probability at least* $1 - (1 - \frac{1}{6^k})^{c6^k}$, *where $k$ is the minimal size of a minimum weight FVS of $G$ and $n$ is the number of vertices.*
*b) The algorithm* SINGLEWGUESSII$(G, k)$ *outputs a feedback vertex set whose expected weight is no more than six times the weight of a minimum weight FVS.*

The proof of each part requires a preliminary lemma.

**Lemma 4** *Let $G = (V, E)$ be a branchy graph, $F$ be a feedback vertex set of $G$ and $X = V \setminus F$. Let $E_X$ denote the set of edges in $E$ whose endpoints are all vertices in $X$ and $E_{F,X}$ denote the set of edges in $G$ that connect vertices in $F$ with vertices in $X$. Then, $|E_X| \leq 2 \cdot |E_{F,X}|$.*

**Proof.** Let $X^b$ be the set of branchpoints in $X$. We replace every linkpoint in $X$ by an edge between its neighbors, and denote the resulting set of edges between vertices in $X^b$ by $E_{X^b}^b$ and between vertices in $X^b$ and $F$ by $E_{F,X^b}^b$. The proof of Lemma 1 shows that

$$|E_{X^b}^b| \leq |E_{F,X^b}^b|.$$

Since every linkpoint in $X$ has both neighbors in the set $X^b \cup F$, the following holds:

$$|E_X| \leq 2 \cdot |E_{X^b}^b| \text{ and } |E_{F,X}| = |E_{F,X^b}^b|.$$

Hence, $|E_X| \leq 2 \cdot |E_{F,X}|$. □

An immediate consequence of Lemma 4 is that the probability of randomly choosing an edge that has at least one endpoint that belongs to a FVS is greater or equal $1/3$. Thus, selecting a vertex at random from a randomly selected edge has a probability of at least $1/6$ to belong to a FVS. Consequently, if the algorithm terminates after $c\,6^k$ iterations, with a WFVS of size $k$, there is a probability of at least $1 - (1 - \frac{1}{6^k})^{c6^k}$ that the output is a minimum WFVS of size at most $k$. This proves part (a) of Theorem 3. Note that since $k$ is not known in advance, we use REPEATEDWGUESSI$(G, c, j)$ with increasing values of $j$ until a FVS is found, say when j=J. When such a set is found it is still possible that there exists a WFVS with more than $J$ vertices that has a smaller weight than the one found. This happens when $k > J$. However, among the WFVSs of size at most $J$, the algorithm finds one with minimum weight with high probability.

The second part requires the following lemma.

**Lemma 5** *Let $G$ be a branchy graph and $F$ be a FVS of $G$. Then,*

$$\sum_{v \in V} d(v) \leq 6 \sum_{v \in F} d(v).$$

**Proof.** Denote by $d_Y(v)$ the number of edges between a vertex $v$ and a set of vertices $Y$. Then,

$$\sum_{v \in V} d(v) = \sum_{v \in X} d(v) + \sum_{v \in F} d(v) =$$
$$\sum_{v \in X} d_X(v) + \sum_{v \in X} d_F(v) + \sum_{v \in F} d(v).$$

Due to Lemma 4,

$$\sum_{v \in X} d_X(v) = 2|E_X| \leq 4|E_{F,X}| = 4 \sum_{v \in X} d_F(v). \tag{3}$$

Consequently,

$$\sum_{v \in V} d(v) \leq 4 \sum_{v \in X} d_F(v) +$$
$$\sum_{v \in X} d_F(v) + \sum_{v \in F} d(v) \leq 6 \sum_{v \in F} d(v)$$

as claimed. $\square$

We can now prove part (b) of Theorem 3 analyzing $\textsc{SingleWGuessII}(G,k)$. Recall that $V_i$ is the set of vertices in graph $G_i$ in iteration $i$, $d_i(v)$ is the degree of vertex $v$ in $G_i$, and $v_i$ is the vertex chosen in iteration $i$. Furthermore, recall that $p_i(v)$ is the probability to choose vertex $v$ in iteration $i$.

The expected weight $E_i(w(v)) = \sum_{v \in V_i} w(v) \cdot p_i(v)$ of a chosen vertex in iteration $i$ is denoted with $a_i$. Thus, due to the linearity of the expectation operator, $E(w(F)) = \sum_{i=1}^{k} a_i$, assuming $|F| = k$. We define a normalization constant for iteration $i$ as follows:

$$\gamma_i = \left[ \sum_{u \in V_i} \frac{d_i(u)}{w(u)} \right]^{-1}$$

Then, $p_i(v) = \gamma_i \cdot \frac{d_i(v)}{w(v)}$ and

$$a_i = \sum_{v \in V_i} w(v) \cdot \frac{d_i(v)}{w(v)} \cdot \gamma_i = \gamma_i \cdot \sum_{v \in V_i} d_i(v)$$

Let $F^*$ be a minimum FVS of $G$ and $F_i^*$ be minimum weight FVS of the graph $G_i$. The expected weight $E_i(w(v)|v \in F_i^*)$ of a vertex chosen from $F_i^*$ in iteration $i$ is denoted with $b_i$. We have,

$$b_i = \sum_{v \in F_i^*} w(v) \cdot p_i(v) = \gamma_i \cdot \sum_{v \in F_i^*} d_i(v)$$

By Lemma 5, $a_i/b_i \leq 6$ for every $i$.

Recall that by definition $F_2^*$ is the minimum FVS in the branchy graph $G_2$ obtained from $G_1 \setminus \{v_1\}$. We get,

$$E(w(F^*)) \geq E_1(w(v)|v \in F_1^*)) + E(w(F_2^*))$$

because the right hand side is the expected weight of the output $F$ assuming the algorithm finds a minimum FVS on $G_2$ and just needs to select one additional vertex, while the left hand side is the unrestricted expectation. By repeating this argument we get,

$$E(w(F^*)) \geq b_1 + E(w(F_2^*)) \geq \cdots \geq \sum_{i=1}^{k} b_i$$

Using $\sum_i a_i / \sum_i b_i \leq \max_i a_i/b_i \leq 6$, we obtain

$$E(w(F)) \leq 6 \cdot E(w(F^*)).$$

Hence, $E(w(F)) \leq 6 \cdot w(F^*)$ as claimed. $\square$

The proof that $\textsc{SingleGuess}(G, k)$ outputs a FVS whose expected size is no more than $4k$ (Theorem 2) where $k$ is the size of a minimum FVS is analogous to the proof of Theorem 3 in the following sense. We assign a weight 1 to all vertices and replace the reference to Lemma 5 by a reference to the following claim: If $F$ is a FVS of a rich graph $G$, then $\sum_{v \in V} d(v) \leq 4 \sum_{v \in F} d(v)$. The proof of this claim is identical to the proof of Lemma 5 except that instead of using Lemma 4 we use Lemma 1.

### 3.3 The Practical Algorithm

In previous sections we presented several algorithms for finding minimum FVS with high probability. The description of these algorithms was geared towards analysis, rather than as a prescription to a programmer. In particular, the number of iterations used within $\textsc{RepeatedWGuessI}(G, c, k)$ is not changed as the algorithm is run with $j < k$. This feature allowed us to regard each call to $\textsc{SingleWGuessI}(G, j)$ made by $\textsc{RepeatedWGuessI}$ as an independent process. Furthermore, there is a small probability for a very long run even when the size of the minimum FVS is small.

We now slightly modify $\textsc{RepeatedWGuessI}$ to obtain an algorithm, termed WRA, which does not suffer from these deficiencies. The new algorithm works as follows. Repeat $\textsc{SingleWGuessI}(G, |V|)$ for $min(Max, c\, 6^{w(F)})$ iterations, where $w(F)$ is the weight of the lightest WFVS found so far and $Max$ is some specified constant determining the maximum number of iterations of $\textsc{SingleWGuessI}$.

### ALGORITHM WRA$(G, c, \textbf{\textit{Max}})$

**Input:** *An undirected weighted graph $G(V, E)$ and constants Max and $c > 1$*
**Output:** *A feedback vertex set $F$*

    $F \leftarrow \textsc{SingleWGuessI}\ (G, |V|)$
    $M \leftarrow min(Max, c\, 6^{w(F)}); i \leftarrow 1;$
    **While** $i \leq M$ **do**
        1. $F' \leftarrow \textsc{SingleWGuessI}(G, |V|)$
        2. **If** $w(F') \leq w(F)$ **then**

| $|V|$ | $|E|$ | values | size | MGA | WRA | Eq. |
|-------|-------|--------|------|-----|-----|-----|
| 15 | 25 | 2–6 | 3–6 | 12 | 81 | 7 |
| 15 | 25 | 2–8 | 3–6 | 7 | 89 | 4 |
| 15 | 25 | 2–10 | 3–6 | 6 | 90 | 4 |
| 25 | 55 | 2–6 | 7–12 | 3 | 95 | 2 |
| 25 | 55 | 2–8 | 7–12 | 3 | 97 | 0 |
| 25 | 55 | 2–10 | 7–12 | 0 | 100 | 0 |
| 55 | 125 | 2–10 | 17–22 | 0 | 100 | 0 |
|  |  |  |  | 31 | 652 | 17 |

Figure 2: Number of graphs in which MGA or WRA yield a smaller loop cutset. The last column records the number of graphs for which the two algorithms produced loop cutsets of the same weight. Each line in the table is based on 100 graphs.

$$F \leftarrow F'; \ M \leftarrow min(Max, c\, 6^{w(F)})$$
3. $i \leftarrow i + 1;$
**End** {While}
Return $F$

**Theorem 6** *If $Max \geq c6^k$, where $k$ is the minimal size of a minimum WFVS of an undirected weighted graph $G$, then WRA($G, c, Max$) outputs a minimum WFVS of $G$ with probability at least $1 - (1 - \frac{1}{6^k})^{c6^k}$.*

The proof is an immediate corollary of Theorem 3.

The choice of $Max$ and $c$ depend on the application. A decision-theoretic approach for selecting such values for any-time algorithms is discussed by Breese and Horvitz (1990).

## 4. Experimental Results

The experiments compared the outputs of WRA vis-à-vis a greedy algorithm GA and a modified greedy algorithm MGA (Becker & Geiger, 1996) based on randomly generated graphs and on some real graphs contributed by the Hugin group (www.hugin.com).

The random graphs are divided into three sets. Graphs with 15 vertices and 25 edges where the number of values associated with each vertex is randomly chosen between 2 and 6, 2 and 8, and between 2 and 10. Graphs with 25 vertices and 55 edges where the number of values associated with each vertex is randomly chosen between 2 and 6, 2 and 8, and between 2 and 10. Graphs with 55 vertices and 125 edges where the number of values associated with each vertex is randomly chosen between 2 and 10. Each instance of the three classes is based on 100 random graphs generated as described by Suermondt and Cooper (1990). The total number of random graphs we used is 700.

The results are summarized in the table of Figure 2. WRA is run with $Max = 300$ and $c = 1$. The two algorithms, MGA and WRA, output loop cutsets of the same size in only

| Name | $|V|$ | $|E|$ | $|F^*|$ | GA | MGA | WRA |
|---|---|---|---|---|---|---|
| Water | 32 | 123 | 16 | 40.7 | 42.7 | 29.5 |
| Mildew | 35 | 80 | 14 | 48.1 | 40.5 | 39.3 |
| Barley | 48 | 126 | 20 | 72.1 | 76.3 | 57.3 |
| Munin1 | 189 | 366 | 59 | 159.4 | 167.5 | 122.6 |

Figure 3: Log size (base 2) of the loop cutsets found by GA, MGA, and WRA.

17 graphs and when the algorithms disagree, then in 95% of these graphs WRA performed better than MGA.

The actual run time of $WRA(G, 1, 300)$ is about 300 times slower than GA (or MGA) on $G$. On the largest random graph we used, it took 4.5 minutes. Most of the time is spend in the last improvement of WRA. Considerable run time can be saved by letting $Max = 5$. For all 700 graphs, WRA(G,1,5) has already obtained a better loop cutset than MGA. The largest improvement, with $Max = 300$, was from a weight of 58.0 ($log_2$ scale) to a weight of 35.9. The improvements in this case were obtained in iterations 1, 2, 36, 83, 189 with respective weights of 46.7, 38.8, 37.5, 37.3, 35.9 and respective sizes of 22, 18, 17, 18, and 17 nodes. On the average, after 300 iterations, the improvement for the larger 100 graphs was from a weight of 52 to 39 and from size 22 to 20. The improvement for the smaller 600 graphs was from a weight of 15 to 12.2 and from size 9 to 6.7.

The second experiment compared between GA, MGA and WRA on four real Bayesian networks showing that WRA outperformed both GA and MGA after a single call to SINGLEWGUESSI. The weight of the output continued to decrease logarithmically with the number of iterations. We report the results with $Max = 1000$ and $c = 1$. Run time was between 3 minutes for Water and 15 minutes for Munin1 on a Pentium 133 with 32M RAM.

## 5. Discussion

Our randomized algorithm, WRA, has been incorporated into the popular genetic software FASTLINK 4.1 by Alejandro Schäffer who develops and maintains this software at the National Institute of Health. WRA replaced previous approximation algorithms for finding FVS because with a small $Max$ value it already matched or improved FASTLINK 4.0 on most datasets examined. The datasets used for comparison are described by Becker et al. (1998). The main characteristics of these datasets is that they were all collected by geneticists, they have a small number of loops, and a large number of values at each node (tens to hundreds depending on the genetic analysis). For such networks the method of conditioning is widely used by geneticists.

The leading inference algorithm, however, for Bayesian networks is the clique-tree algorithm (Lauritzen & Spiegelhalter, 1988) which has been further developed in several papers (Jensen, Lauritzen, & Olsen, 1990a; Jensen, Olsen, & Andersen, 1990b). For the networks presented in Table 3 conditioning is not a feasible method while the clique tree algorithm can and is being used to compute posterior probabilities in these networks. Furthermore, it has been shown that the weight of the largest clique is bounded by the weight of the loop cutset union the largest parent set of a vertex in a Bayesian network implying that the clique tree algorithm is always superior in time performance over the conditioning algorithm (Shachter, Andersen, & Szolovits, 1994). The two methods, however, can be combined to strike a balance between time and space requirements as done within the bucket elimination framework (Dechter, 1999).

The algorithmic ideas behind the randomized algorithms presented herein can also be applied for constructing good clique trees and initial experiments confirm that an improvement over deterministic algorithms is often obtained. The idea is that instead of greedily selecting the smallest clique when constructing a clique tree, one would randomly select the next clique according to the relative weights of the candidate cliques. It remains to develop the theory behind random choices of clique trees before a solid assessment can be presented. Currently, there is no algorithm for finding a clique tree such that its size is guaranteed to be close to optimal with high probability.

Horvitz et al. (1989) show that the method of conditioning can be useful for approximate inference. In particular, they show how to rank the instances of a loop cutset according to their prior probabilities assuming all variables in the cutset are marginally independent. The conditioning algorithm can then be run according to this ranking and the answer to a query be given as an interval that shrinks towards the exact solution as more instances of the loop cutset are considered (Horvitz, Suermondt, & Cooper, 1989; Horvitz, 1990). Applying this idea without making independence assumptions is described by Darwiche (1994). So if the maximal clique is too large to store one can still perform approximate inferences using the conditioning algorithm.

## Acknowledgment

We thank Seffi Naor for fruitful discussions. Part of this work was done while the third author was on sabbatical at Microsoft Research. A variant of this work has been presented at the fifteenth conference on uncertainty in artificial intelligence, July 1999, Sweden.